\documentclass[11pt,a4paper]{article}
\usepackage[nohyperref]{acl2017}
\usepackage{times}
\usepackage{latexsym}

\usepackage{url}

\usepackage{tikz}
\usepackage{times}
\usepackage{graphicx}
\usepackage{latexsym}
\usepackage{amsmath}
\usepackage{amssymb}
\usepackage{multirow}
\usepackage{color}
\usepackage{tablefootnote}

\usepackage{booktabs}
\usepackage{url}
\usepackage[font={footnotesize}]{caption}

\newcommand{\comment}[1]{}
\aclfinalcopy 

\DeclareMathOperator*{\argmax}{arg\,max}

\title{Coarse-to-Fine Question Answering for Long Documents}
  
  \author{Eunsol Choi \\ University of Washington\\{\tt eunsol@cs.washington.edu}
   \And Daniel Hewlett, Alexandre Lacoste\\ Google Research \\ {\tt \{dhewlett,allac\}@google.com}  \\
         \AND
       Illia Polosukhin , Jakob Uszkoreit\\ Google Research \\ {\tt \{ipolosukhin,usz\}@google.com} 
               \And
         Jonathan Berant \\  Tel Aviv University \\  {\tt 
 joberant@cs.tau.ac.il}}
\date{}

\begin{document}
\maketitle
\begin{abstract}
%
{
We present a framework for question answering that can efficiently scale to longer documents while maintaining or even improving performance of state-of-the-art models. While most successful approaches for reading comprehension rely on recurrent neural networks (RNNs), running them over long documents is prohibitively slow because it is difficult to parallelize over sequences.
Inspired by how people first skim the document, identify relevant parts, and carefully read these parts to produce an answer, we combine a coarse, fast model for selecting relevant sentences and a more expensive RNN for producing the answer from those sentences.
We treat sentence selection as a latent variable trained jointly from the answer only using reinforcement learning.
Experiments demonstrate the state of the art performance on a challenging subset of the \textsc{Wikireading} dataset \cite{Hewlett:16} and on a new dataset, while speeding up the model by 3.5x-6.7x.
}

\end{abstract}

\section{Introduction}
Reading a document and answering questions about its content are among the hallmarks of natural language understanding. 
\begin{figure}
\centering
\usetikzlibrary{shapes.geometric}

\vspace{-7pt}
 \footnotesize
\centering
\tikzstyle{edgeB} = [draw,thick,->,blue,dashed]
\tikzstyle{edgeBL} = [draw,line width=1.5pt,->,black]
\tikzstyle{edgeR} = [draw,thick,->,black,dashed]
\tikzstyle{edgeRB} = [draw,line width=1.5pt,->,red]
\tikzstyle{edgeB} = [draw,thick,->,blue,loosely dashed]
\tikzstyle{edgeBB} = [draw,line width=1.5pt,->,blue]
\tikzstyle{vertex} = [ellipse,draw]%
\tikzstyle{vertex2} = [rectangle,draw=black]%
\tikzstyle{model} = [rectangle,fill=black!10,minimum size=20pt]%
\begin{tikzpicture}[scale=1.5,auto,swap]
  \node[vertex]      (q)    at (-0.7,3.6)    {Query ($x$)};
    \node[vertex]        (d)  at (2.7,3.6)  {Document ($d$)};
    \node[vertex2]       (a)   at (1.0,0.8) {Answer ($y$)};
    \node[model]       (smodel)   at (1,3.0) {Sentence Selection (Latent)};
    \node[model]       (amodel)   at (1,1.4) {Answer Generation (RNN)};
    \node[vertex2]       (sub)  at (1,2.2)  {Document Summary ($\hat{d})$};
  \path[edgeB] (q) -- (smodel);
  \path[edgeB] (d) -- (smodel);
  \draw[->,blue,dashed,thick] (q.200) -- ++(south:1.5cm)  --(amodel.west);
 \path[edgeR] (smodel) -- (sub);
 \path[edgeB] (sub) -- (amodel);
 \path[edgeR] (amodel) -- (a);
\end{tikzpicture}
\caption{\small Hierarchical question answering: the model first selects relevant sentences that produce a document summary ($\hat{d}$) for the given query ($x$), and then generates an answer ($y$) based on the summary ($\hat{d}$) and the query $x$.} \vspace{-15pt}
\label{fig:overview}
\end{figure}
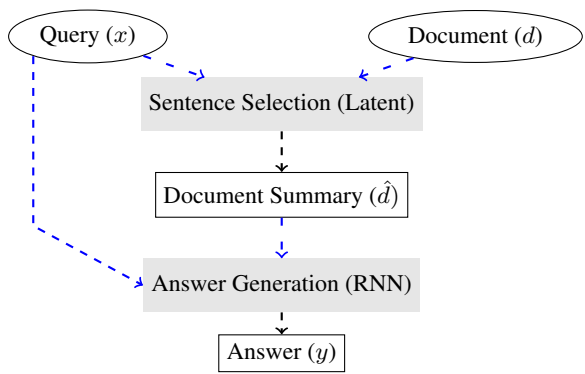
Recently, interest in question answering (QA) from unstructured documents has increased along with the availability of large scale datasets for reading comprehension \cite{Hermann:15,Hill:15,Rajpurkar:16,Onishi:16,Nguyen:16,Trischler:16}.

Current state-of-the-art approaches for QA over documents are based on recurrent neural networks (RNNs) that
encode the document and the question to determine the answer~\cite{Hermann:15,chen2016thorough,Kumar:16,Kadlec:16,Xiong:16}.
While such models have access to all the relevant information, they are slow because the model needs to be run sequentially over possibly thousands of tokens, and the computation is not parallelizable.
In fact, such models usually truncate the documents and
consider only a limited number of tokens~\cite{Miller:16,Hewlett:16}.
Inspired by studies on how people answer questions by first skimming the document, identifying relevant parts, and carefully reading these parts to produce an answer~\cite{Masson:83}, we propose a coarse-to-fine model for question answering.

Our model takes a hierarchical approach (see Figure~\ref{fig:overview}), where first a fast
model is used to select
a few sentences from the document that are relevant for
answering the question~\cite{Yu:14,WikiQA:16}. Then, a slow RNN is employed to produce the final answer from the selected sentences. The RNN is run over a fixed number of tokens, regardless of the length of the document. Empirically, our model encodes the text up to 6.7 times faster than the base model, which reads the first few paragraphs, while having access to four times more tokens.

A defining characteristic of our setup is that an answer does not necessarily appear verbatim in the input (the genre of a movie can be
determined even if not mentioned explicitly). Furthermore, the answer often appears multiple times in the document in spurious contexts (the year `2012' can appear many times while only once in relation to the question). Thus, we treat sentence selection as a latent variable that is trained jointly with the answer generation model from the answer only using reinforcement learning. Treating sentence selection as a latent variable has been explored in classification~\cite{yessenalina2010multi,Lei:16}, however, to our knowledge, has not been applied for question answering.

We find that jointly training sentence selection and answer generation is especially helpful when locating the sentence containing the answer is hard. 
We evaluate our model
on the \textsc{Wikireading} dataset~\cite{Hewlett:16}, 
focusing on examples where the document is long and sentence selection is challenging, and on a new dataset called \textsc{Wikisuggest} that contains more natural questions gathered from a search engine.

To conclude, we present a modular framework and learning procedure for QA over long text. 
It captures a limited form of document structure such as sentence boundaries 
and deals with long documents or potentially multiple documents. 
Experiments show improved performance compared to the state of the art on the subset of \textsc{Wikireading}, comparable performance on other datasets, and a 3.5x-6.7x speed up in document encoding, while allowing access to much longer documents.

\begin{figure}
\centering
\noindent
\small 

$d$:
\fbox{\begin{minipage}{21em}
$s_1$: The 2011 Joplin tornado was a catastrophic EF5-rated multiple-vortex tornado that struck Joplin, Missouri $\ldots$

$s_4$: It was the third tornado to strike Joplin since May 1971.

$s_5$: Overall, the tornado killed \textbf{158 people} $\ldots$, injured some 1,150 others, and caused damages $\ldots$
\end{minipage}}
\begin{minipage}{23em}
\noindent \textbf{$x$}: how many people died in joplin mo tornado\\
\noindent \textbf{$y$}: 158 people
\end{minipage}
\caption{\small A training example containing a document $d$, a question $x$ and an answer $y$ in the \textsc{WikiSuggest} dataset. In this example, the sentence $s_5$ is necessary to answer the question.}
\vspace{-8pt}
\label{fig:factoid}
\end{figure}
\section{Problem Setting}
\vspace{-4pt}
Given a training set of question-document-answer triples $\{x^{(i)}, d^{(i)}, y^{(i)}\}_{i=1}^N$, our goal is to learn a model that produces an answer $y$ for a question-document pair $(x,d)$. A document $d$ is a list of sentences $s_1, s_2, \dots, s_{|d|}$, and we assume that the answer can be produced from a small latent subset of the sentences. Figure~\ref{fig:factoid} illustrates a training example in which sentence $s_5$ is in this subset.
\vspace{-4pt}
\section{Data}\label{sec:data}
\vspace{-4pt}
We evaluate on \textsc{WikiReading}, \textsc{WikiReading Long}, and a new dataset, \textsc{WikiSuggest}. 

\textsc{Wikireading} \cite{Hewlett:16} is a QA dataset automatically generated from Wikipedia and Wikidata: given a Wikipedia page about an entity and a Wikidata property, such as \textsc{profession}, or \textsc{gender}, the goal is to infer the target value based on the document. Unlike other recently released large-scale datasets~\cite{Rajpurkar:16,Trischler:16}, \textsc{Wikireading} does not annotate answer spans, making sentence selection more challenging. 

Due to the structure and short length of most Wikipedia documents (median number of sentences: 9), the answer can usually be inferred from the first few sentences. Thus, the data is not ideal for testing a sentence selection model compared to a model that uses the first few sentences. 
Table~\ref{tab:answerMatch} quantifies this intuition: We consider sentences containing the answer $y^*$ as a proxy for sentences that should be selected, and report how often $y^*$ appears in the document. Additionally, we report how frequently this proxy oracle sentence is the first sentence. We observe that in \textsc{Wikireading}, the answer appears verbatim in 47.1\% of the examples, and in 75\% of them 
the match is in the first sentence. Thus, the importance of modeling sentence selection is limited. 

\begin{table}
\begin{center}
\scriptsize{
\begin{tabular}{c|c|c|c}
 \toprule
 & \% answer & avg \# of & \% match \\ 
 & string exists & answer match & first sent\\ \midrule
\textsc{Wikireading}       & 47.1 & 1.22   & 75.1 \\ 
\textsc{WR-Long}&{50.4}& {2.18} &{31.3}\\ 
\textsc{Wikisuggest}          & 100  & 13.95   & 33.6 \\ \toprule
\end{tabular}}
\end{center}
\vspace{-8pt}
\caption{\small{Statistics on string matches of the answer $y^*$ in the document. The third column only considers examples with answer match. Often the answer string is missing or appears many times while it is relevant to query only once.}}
\vspace{-5pt}
\label{tab:answerMatch}
\end{table}

\begin{table}
\scriptsize{
\begin{center}
\begin{tabular}{c|c|c|c|c}
 \toprule
             &\# of uniq.& \# of  & \# of words&\# of tokens \\
            & queries & examples & / query&  / doc.\\ \midrule 
 \textsc{Wikireading}            & 858              & 16.03M         & 2.35                   & 568.9 \\
 \textsc{WR-Long} & 239              & 1.97M          & 2.14                     &1200.7               \\ 
 \textsc{Wikisuggest}           & 3.47M            & 3.47M          & 5.03       & 5962.2 \\ \toprule 
\end{tabular}
\end{center}
\vspace{-10pt}
\caption{\small Data statistics.}
\vspace{-8pt}
\label{tab:datastat}}
\end{table}
To remedy that, we filter \textsc{Wikireading} and ensure a more even distribution of answers throughout the document. We prune short documents with less than 10 sentences, and only consider Wikidata properties for which \newcite{Hewlett:16}'s best model obtains an accuracy of less than 60\%. This prunes out properties such as \textsc{Gender}, \textsc{Given Name}, and \textsc{Instance Of}.\footnote{These three relations alone account for 33\% of the data.}
The resulting \textsc{Wikireading Long} dataset contains 1.97M examples, where the answer appears in 50.4\% of the examples, and appears in the first sentence only 31\% of the time. On average, the documents in \textsc{Wikireading Long} contain 1.2k tokens, more tokens than those of SQuAD (average 122 tokens) or CNN (average 763 tokens) datasets (see Table~\ref{tab:datastat}). Table~\ref{tab:answerMatch} shows that the exact answer string is often missing from the document in \textsc{Wikireading}. This is since Wikidata statements include
properties such as \textsc{Nationality}, which are not explicitly mentioned, but can still be inferred. A drawback of this dataset is that the queries, Wikidata properties, are not natural language questions and are limited to 858 properties.

\begin{figure}
\centering
\includegraphics[scale=0.16]{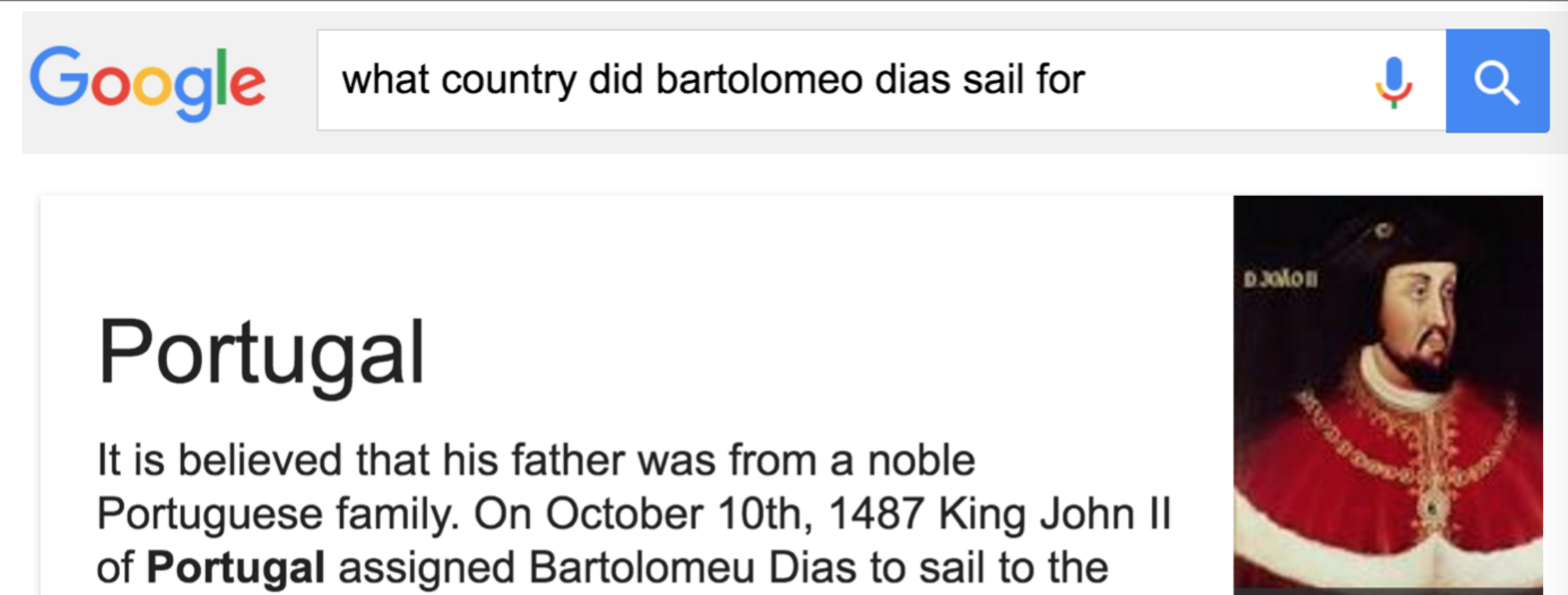}
\scriptsize{
\begin{tabular}{l|l}
\multicolumn{2}{}{}\\\midrule
\textsc{WikiSuggest} Query  & Answer  \\ \midrule
what year did virgina became a state& 1788\\
general manager of smackdown& Theodore Long\\
minnesota viking colors& purple\\
coco martin latest movies& maybe this time\\
longest railway station in asia&Gorakhpur\\
son from modern family & Claire Dunphy\\
north dakota main religion & Christian\\
lands end' brand	 & Lands' End\\
wdsu radio station& WCBE\\
\toprule
\end{tabular}
}
\vspace{-5pt}
\caption{Example queries and answers of \textsc{Wikisuggest}.}
\vspace{-10pt}
\label{fig:visual_query}
\end{figure}

To model more realistic language queries, we collect the \textsc{Wikisuggest} dataset as follows. We use the Google Suggest API to harvest natural language questions 
and submit them to Google Search.
Whenever Google Search returns a box with a short answer from Wikipedia (Figure ~\ref{fig:visual_query}), we create an example from the question, answer, and the Wikipedia document. 
If the answer string is missing from the document this often implies a spurious question-answer pair, such as (`what time is half time in rugby', `80 minutes, 40 minutes'). Thus, we pruned question-answer pairs without the exact answer string. 
We examined fifty examples after filtering and found that 54\% were well-formed question-answer pairs where we can ground answers in the  document, 20\% contained answers without textual evidence in the document (the answer string exists in an irreleveant context), and 26\% contain incorrect QA pairs such as the last two examples in Figure~\ref{fig:visual_query}.

\vspace{-4pt}
\section{Model}\label{sec:model}
\vspace{-4pt}
Our model has two parts (Figure~\ref{fig:overview}): a fast sentence selection model (Section~\ref{subsec:sent_select}) that defines a distribution $p(s \mid x, d)$ over sentences given the input question ($x$) and the document ($d$), and a more costly answer generation model (Section~\ref{subsec:answer_gen}) that generates an answer $y$ 
given the question and a document summary, $\hat{d}$ (Section \ref{subsec:doc_summary}), that focuses on the relevant parts of the document. 

\subsection{Sentence Selection Model} \label{subsec:sent_select}
Following recent work on sentence selection~\cite{Yu:14,Yang:16}, we build a feed-forward network to define a distribution over the sentences $s_1, s_2, \dots, s_{|d|}$.
We consider three simple sentence representations: a bag-of-words (BoW) model, a chunking model, and a (parallelizable) convolutional model. These models are efficient at dealing with long documents, but do not fully capture the sequential nature of text. 

\paragraph{BoW Model}
Given a sentence $s$, we denote by $\text{BoW}(s)$ the bag-of-words representation that averages the embeddings of the tokens in $s$.
To define a distribution over the document sentences, we employ a standard attention model (e.g., \cite{Hermann:15}), where the BoW representation of the query is concatenated to the BoW representation of each sentence $s_l$, and then passed through a single layer feed-forward network:
\begin{equation*} 
\begin{split}
h_l & =[\text{BoW}(x);\text{BoW}(s_l)] \\
v_l & = v^\top \text{ReLU}(W  h_l), \\
p(s=s_l \mid x, d) & = \text{softmax}(v_l),
\end{split}
\end{equation*}
where  $[;]$ indicates row-wise concatenation, and the matrix $W$, the vector $v$, and the word embeddings are learned parameters.

\paragraph{Chunked BoW Model}
To get more fine-grained granularity, we split sentences into fixed-size smaller chunks (seven tokens per chunk) and score each chunk separately \cite{Miller:16}. This is beneficial if questions are answered with sub-sentential units, by allowing to learn attention over different chunks. 
We split a sentence $s_l$ into a fixed number of chunks ($c_{l,1},c_{l,2}\ldots,c_{l,J}$), generate a BoW representation for each chunk, and score it exactly as in the BoW model. We obtain a distribution over chunks, and compute sentence probabilities by marginalizing over chunks from the same sentence. Let $p(c=c_{l,j} \mid x, d)$ be the distribution over chunks from all sentences, then:
\begin{equation*} 
\begin{split}
p(s=s_l \mid x, d) = \sum_{j=1}^{J} p(c=c_{l,j} \mid x, d), 
\end{split}
\end{equation*}
with the same parameters as in the BoW model.
\paragraph{Convolutional Neural Network Model}
While our sentence selection model is designed to be fast, we explore a convolutional neural network (CNN) that can compose the meaning of nearby words. A CNN is still efficient, since all filters can be computed in parallel.
Following previous work~\cite{Kim:14,Kalchbrenner:14}, we concatenate the embeddings of tokens in the query $x$ and the sentence $s_l$, and run a convolutional layer with $F$ filters and width $w$ over the concatenated embeddings. This results in $F$ features for every span of length $w$, and we employ max-over-time-pooling \cite{collobert11scratch} 
to get a final representation $h_l \in \mathbb{R}^F$. We then compute $p(s=s_l \mid x, d)$ by passing $h_l$ through a single layer feed-forward network as in the BoW model.



\subsection{Document Summary} \label{subsec:doc_summary}
After computing attention over sentences, we create a summary that focuses on the document parts related to the question using deterministic soft attention or stochastic hard attention. Hard attention is more flexible, as it can focus on multiple sentences, while soft attention is easier to optimize and retains information from multiple sentences.

\paragraph{Hard Attention}
We sample a sentence $\hat{s} \sim p(s \mid x, d )$ and fix the document summary $\hat{d}=\hat{s}$ to be that sentence during training. At test time, we choose the most probable sentence. To extend the document summary to contain more information, we can sample without replacement $K$ sentences from the document and define the summary to be the concatenation of the sampled sentences $\hat{d} = [\hat{s}_1 ;\hat{s}_2 ; \dots ;  \hat{s}_K]$.




\paragraph{Soft Attention}
In the soft attention model~\cite{Bahdanau:14} we compute a weighted average of the tokens in the sentences according to $p(s \mid x, d)$. More explicitly, let $\hat{d}_m$ be the $m$th token of the document summary. Then, by fixing the length of every sentence to $M$ tokens,\footnote{Long sentences are truncated and short ones are padded.} the \textit{blended} tokens are computed as follows:\vspace{-5pt}
\begin{align*}
\hat{d}_m = \sum_{l=1}^{|d|} p(s = s_l \mid x, d) \cdot s_{l,m},
\end{align*}\vspace{-5pt}
where $s_{l,m}$ is the $m$th word in the $l$th sentence ($m \in \{1, \dots, M\}$).

As the answer generation models (Section~\ref{subsec:answer_gen}) take a sequence of vectors as input, we average the tokens at the word level. This gives the hard attention an advantage since it samples a ``real" sentence without mixing words from different sentences. Conversely, soft attention is trained more easily, and has the capacity to learn a low-entropy distribution that is similar to hard attention.
\vspace{-5pt}
\subsection{Answer Generation Model} \label{subsec:answer_gen}
State-of-the-art question answering models~\cite{chen2016thorough,Seo:16} use RNN models to encode the document and question and selects the answer. We focus on a hierarchical model with fast sentence selection, and do not subscribe to a particular answer generation architecture. 

Here we implemented the state-of-the-art word-level sequence-to-sequence model with placeholders, described by \newcite{Hewlett:16}. This models can produce answers that does not appear in the sentence verbatim. 
This model takes the query tokens, and the document (or document summary) tokens as input and encodes them with a Gated Recurrent Unit (GRU;~\newcite{Cho:14}). Then, the answer is decoded 
with another GRU model, defining a distribution over answers $p(y \mid x, \hat{d})$. In this work, we modified the original RNN: the word embeddings for the RNN decoder input, output and original word embeddings are shared. 

\section{Learning}\label{sec:learning}
We consider three approaches for learning the model parameters (denoted by $\theta$): (1) We present a pipeline model, where we use distant supervision to train a sentence selection model independently from an answer generation model. (2) The hard attention model is optimized with REINFORCE~\cite{Williams:92} algorithm. (3) The soft attention model is fully differentiable and is optimized end-to-end with backpropagation.

\paragraph{Distant Supervision}
While we do not have an explicit supervision for sentence selection, we can define a simple heuristic for labeling sentences. We define the gold sentence to be the first sentence that has a full match of the answer string, or the first sentence in the document if no full match exists.
By labeling gold sentences, we can train sentence selection and answer generation independently with standard supervised learning, maximizing the log-likelihood of the gold sentence and answer, given the document and query.
Let $y^*$ and $s^*$ be the target answer and sentence , where $s^*$ also serves as the document summary. The objective is to maximize:
\begin{align*} 
J(\theta) &= \log p_\theta(y^*, s^* \mid x, d) \\
 &= \log p_\theta (s^* \mid x,d) + \log p_\theta (y^* \mid s^*,x).
\end{align*}
Since at test time we do not have access to the target sentence $s^*$ needed for answer generation, we replace it by the model prediction $\argmax_{s_l \in d} p_\theta(s=s_l\mid d,x)$.

\paragraph{Reinforcement Learning}
Because the target sentence is missing, we use reinforcement learning 
where our action is sentence selection, and our goal is to select sentences that lead to a high reward.
We define the reward for selecting a sentence as the log probability of the correct answer given that sentence, that is, 
$R_\theta(s_l) = \log p_\theta(y\! =\! y^* \mid s_l, x)$. Then the learning objective is to maximize the expected reward: 
\begin{align*}
J(\theta) =& \sum_{s_l \in d} p_\theta(s\! =\! s_l \mid x, d) \cdot  R_\theta(s_l) \\
=& \sum_{s_l \in d}  p_\theta(s\!=\! s_l \mid x, d ) \cdot \log p_\theta(y\! =\! y^* \mid s_l, x).
\end{align*}

Following REINFORCE~\cite{Williams:92}, we approximate the gradient of the objective with a sample, $\hat{s} \sim p_\theta(s \mid x, d)$:
\begin{align*}
\nabla J(\theta) \approx & \; \nabla \log p_\theta(y \mid \hat{s}, x) \\
&+ \log p_\theta(y \mid \hat{s}, x) \cdot \nabla \log p_\theta(\hat{s} \mid x, d).
\end{align*}
Sampling $K$ sentences is similar and omitted for brevity.

Training with REINFORCE is known to be unstable due to the high variance induced by sampling. To reduce variance, we use curriculum learning, start training with distant supervision and gently transition to reinforcement learning, similar to \textsc{DAgger}~\cite{Ross:11}. Given an example, we define the probability of using the distant supervision objective at each step as $r^{e}$, where $r$ is the decay rate and $e$ is the index of the current training epoch.\footnote{ We tuned $r \in [0.3, 1]$ on the development set.}

\paragraph{Soft Attention}
We train the soft attention model by maximizing the log likelihood of the correct answer $y^*$ given the input question and document $\log p_\theta (y^*\ |\ d, x)$. Recall that the answer generation model takes as input the query $x$ and document summary $\hat{d}$, and since $\hat{d}$ is an average of sentences weighted by sentence selection, the objective is differentiable and is trained end-to-end.

\section{Experiments}

\paragraph{Experimental Setup}
We used 70\% of the data for training, 10\% for development, and 20\% for testing in all datasets.
We used the first 35 sentences in each document as input to the hierarchical models, where each sentence has a maximum length of 35 tokens. Similar to \newcite{Miller:16}, we add the first five words in the document (typically the title) at the end of each sentence sequence for \textsc{WikiSuggest}. We add the sentence index as a one hot vector to the sentence representation. We coarsely tuned and fixed most hyper-parameters for all models, and separately tuned the learning rate and gradient clipping coefficients for each model on the development set.  The details are reported in the supplementary material.

\paragraph{Evaluation Metrics}
Our main evaluation metric is answer accuracy, the proportion of questions answered correctly. 
For sentence selection, since we do not know which sentence contains the answer, we report approximate sentence selection accuracy by matching sentences that contain the answer string ($y^*$). For the soft attention model, we treat the sentence with the highest probability as the predicted sentence. 

\paragraph{Models and Baselines}
The models \textsc{Pipeline}, \textsc{Reinforce}, and \textsc{SoftAttend} correspond to the learning objectives in Section~\ref{sec:learning}. We compare these models against the following baselines:
\begin{itemize}
  \setlength{\itemsep}{0pt}
  \setlength{\parskip}{0pt}
  \setlength{\parsep}{0pt}  
  \setlength{\topsep}{0pt}
    \item[] \textbf{\textsc{First}} always selects the first sentence of the document. The answer appears in the first sentence in 33\% and 15\% of documents in \textsc{Wikisuggest} and \textsc{Wikireading Long}.
     \item[] \textbf{\textsc{Base}} is the re-implementation of the best model by \newcite{Hewlett:16}, consuming the first 300 tokens. We experimented with providing additional tokens to match the length of document available to hierarchical models, but this performed poorly.\footnote{Our numbers on \textsc{Wikireading} outperform previously reported numbers due to modifications in implementation and better optimization.}
    \item[] \textbf{\textsc{Oracle}} selects the first sentence with the answer string if it exists, or otherwise the first sentence in the document.
\end{itemize}
\begin{table}
\small
\begin{center}
\begin{tabular}{c|c|c}
 \hline
Dataset & Learning &Accuracy\\ \hline
 &  \textsc{First} &  26.7 \\
 &  \textsc{Base} & 40.1 \\
&  \textsc{Oracle} & 43.9 \\  \cline{2-3}
\textsc{Wikireading} &  \textsc{Pipeline}   & 36.8 \\
\textsc{Long} & \textsc{SoftAttend} & 38.3 \\
  &  \textsc{Reinforce} ($K$=1)  & 40.1\\
&\textsc{Reinforce} ($K$=2)  & \textbf{42.2} \\ \hline
 &  \textsc{First} &  44.0 \\
 &  \textsc{Base} &  \bf{46.7} \\
 &  \textsc{Oracle} &  60.0\\ \cline{2-3}
\textsc{Wiki} &  \textsc{Pipeline}   & 45.3 \\ 
\textsc{suggest}  & \textsc{SoftAttend} &  45.4 \\
&  \textsc{Reinforce} ($K$=1) & 45.4 \\
 &  \textsc{Reinforce} ($K$=2)&  45.8 \\
 \hline
 &  \textsc{First} & 71.0 \\
 &  \textsc{Hewlett et al. (2016)} & 71.8 \\
 &  \textsc{Base} & \textbf{75.6} \\ 
 &  \textsc{Oracle} & 74.6  \\  
 \cline{2-3}
\textsc{Wikireading}   & \textsc{SoftAttend} & 71.6 \\
 &  \textsc{Pipeline}  & 72.4  \\
 &  \textsc{Reinforce} ($K$=1)  & 73.0  \\
&\textsc{Reinforce} ($K$=2)  & 74.5\\\cline{2-3}
\cline{2-3}
 \hline
\end{tabular}
\end{center}
\vspace{-6pt}
\caption{\small{Answer prediction accuracy on the test set. $K$ is the number of sentences in the document summary.}}
\vspace{-8pt}
\label{tab:ans_result}
\end{table}

\begin{figure}
\centering
\includegraphics[scale=0.45]{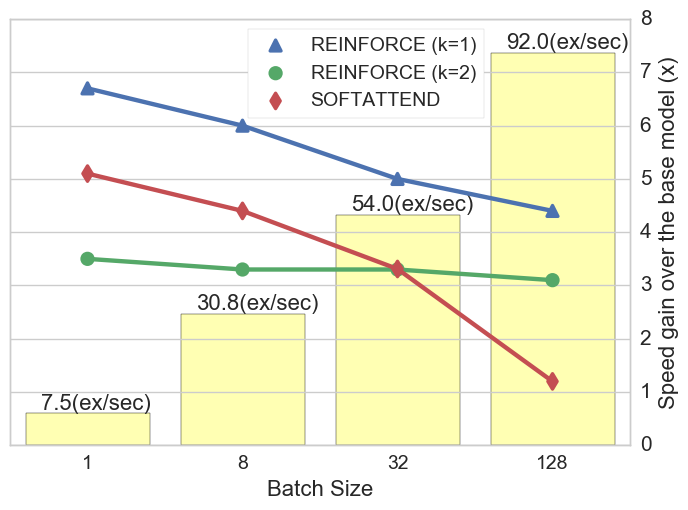}
\vspace{-22pt}
\caption{\small{Runtime for document encoding on an Intel Xeon CPU E5-1650 @3.20GHz on \textsc{Wikireading} at test time. The boxplot represents the throughput of \textsc{Base} and each line plot shows the proposed models' speed gain over \textsc{Base}. 
Exact numbers are reported in the supplementary material.}}\label{fig:speed}
\vspace{-10pt}
\end{figure}
%
\paragraph{Answer Accuracy Results}
Table~\ref{tab:ans_result} summarizes answer accuracy on all datasets. We use \textsc{Bow} encoder for sentence selection as it is the fastest. 
The proposed hierarchical models match or exceed the performance of \textsc{Base}, while reducing the number of RNN steps significantly, from 300 to 35 (or 70 for $K$=2), and allowing access to later parts of the document.
Figure~\ref{fig:speed} reports the speed gain of our system. 
While throughput at training time can be improved by increasing the batch size,
at test time real-life QA systems use batch size $1$, where \textsc{Reinforce} obtains a 3.5x-6.7x speedup (for $K$=2 or $K$=1). In all settings, \textsc{Reinforce} was at least three times faster than the \textsc{Base} model.

All models outperform the \textsc{First} baseline, 
and utilizing the proxy oracle sentence (\textsc{Oracle}) improves performance on \textsc{Wikisuggest} and \textsc{Wikireadng Long}. In \textsc{Wikireading}, where the proxy oracle sentence is often missing and documents are short, \textsc{Base} outperforms \textsc{Oracle}.

Jointly learning answer generation and sentence selection, \textsc{Reinforce} 
outperforms \textsc{Pipeline}, which relies on a noisy supervision signal for sentence selection. The improvement is larger in \textsc{Wikireading Long}, where the approximate supervision for sentence selection is missing for 51\% of examples compared to 22\% of examples in \textsc{Wikisuggest}.\footnote{The number is lower than in Table~\ref{tab:answerMatch} because we cropped sentences and documents, as mentioned above.}

On \textsc{Wikireading Long}, \textsc{Reinforce} outperforms all other models (excluding \textsc{Oracle}, which has access to gold labels at test time). 
In other datasets, \textsc{Base} performs slightly better than the proposed models, at the cost of speed. In these datasets, the answers are concentrated in the first few sentences. \textsc{Base} is advantageous in categorical questions (such as \textsc{Gender}), gathering bits of evidence from the whole document, at the cost of speed. Encouragingly, our system almost reaches the performance of \textsc{Oracle} in \textsc{Wikireading}, showing strong results in a limited token setting.

Sampling an additional sentence into the document summary increased performance in all datasets, illustrating the flexibility of hard attention compared to soft attention. 
Additional sampling allows recovery from mistakes in \textsc{Wikireading Long}, where sentence selection is challenging.\footnote{Sampling more help pipeline methods less.} Comparing hard attention to soft attention, we observe that \textsc{Reinforce} performed better than \textsc{SoftAttend}.
The attention distribution learned by the soft attention model was often less peaked, generating noisier summaries.\footnote{We provide a visualization of the attention distribution for different learning methods in the supplementary material.}


\paragraph{Sentence Selection Results}

\begin{table}
\begin{center}
\footnotesize{
\begin{tabular}{c|c|c|c}
\hline
Dataset     & Learning  & Model     &    Accuracy    \\ \hline
            &           & \textsc{CNN}       & 70.7  \\
            & \textsc{Pipeline}  & \textsc{BoW}       & 69.2  \\ 
            &           & \textsc{ChunkBoW}  & \bf{74.6}  \\ \cline{2-4}
          \textsc{Wiki}  &           & \textsc{CNN}       & 74.2 \\
\textsc{reading}          & \scriptsize{\textsc{Reinforce}} & \textsc{BoW} & 72.2 \\
\textsc{Long}    &           & \textsc{ChunkBoW}  & \textbf{74.4}  \\ \cline{2-4} 
            & \multicolumn{2}{|c|}{\textsc{First}}& 31.3 \\
            & \multicolumn{2}{|c|}{\textsc{SoftAttend} (BoW)} & 70.1 \\
 \hline
            &           & \textsc{CNN}       & 62.3  \\
       & \textsc{Pipeline}  & \textsc{BoW}       & \bf{67.5} \\ 
            &           & \textsc{ChunkBoW}  & 57.4 \\ \cline{2-4}
       \textsc{Wiki}       &           & CNN       & 64.6  \\
   \textsc{suggest} & \scriptsize{\textsc{Reinforce}} & \textsc{BoW} & \textbf{67.3} \\
   &   & \textsc{ChunkBoW}  & 59.3 \\ \cline{2-4}
            & \multicolumn{2}{|c|}{\textsc{First}}  & 42.6 \\
            & \multicolumn{2}{|c|}{\textsc{SoftAttend} (BoW)}  & 49.9 \\\hline
\end{tabular}}
\end{center}
\vspace{-8pt}
\caption{Approximate sentence selection accuracy on the development set for all models. We use \textsc{Oracle} to find a proxy gold sentence and report the proportion of times each model selects the proxy sentence.}
\vspace{-14pt}
\label{tab:sent_result} 
\end{table}

Table~\ref{tab:sent_result} reports sentence selection accuracy by showing the proportion of times models selects the proxy gold sentence when it is found by \textsc{Oracle}. 
In \textsc{Wikireading Long}, \textsc{Reinforce} finds the approximate gold sentence in 74.4\% of the examples where the the answer is in the document. In \textsc{Wikisuggest} performance is at 67.5\%, mostly due to noise in the data.
\textsc{Pipeline} performs slightly better as it is directly trained towards our noisy evaluation. However, not all sentences that contain the answer are useful to answer the question (first example in Table~\ref{tab:full_examples}). \textsc{Reinforce} learned to choose sentences that are likely to generate a correct answer rather than proxy gold sentences, improving the final answer accuracy. 
On \textsc{Wikireading Long}, complex models (\textsc{CNN} and \textsc{ChunkBow}) outperform the simple \textsc{BoW}, while on \textsc{Wikisuggest} \textsc{Bow} performed best.
\begin{table*}[t!]
\footnotesize{
\begin{tabular}{c|c|c|l|}\cmidrule[1pt]{2-4}
\parbox[t]{2mm}{\multirow{15}{*}{\rotatebox[origin=c]{90}{\textsc{WikiReading Long (WR Long)}}}}&

\multicolumn{2}{|c|}{Error Type} &No evidence in doc. \\
&\multicolumn{2}{|c|}{(Query, Answer)} &(place\_of\_death, Saint Petersburg)\\
&\multicolumn{2}{|c|}{System Output} &Crimean Peninsula \\  \cmidrule[0.5pt]{2-4}
&1 &  11.7&Alexandrovich Friedmann ( also spelled Friedman or [Fridman] , Russian :  $\ldots$\\
&4 &3.4 &Friedmann was baptized {\ldots} 
and lived much of his life in Saint Petersburg . \\
&25 &\bf{63.6} &Friedmann died on September 16 , 1925 , at the age of 37 , from typhoid fever that \\
&& &he contracted while returning from a vacation in {Crimean Peninsula} .  \\
 \cmidrule[1.0pt]{2-4}
 &\multicolumn{2}{|c|}{Error Type} & Error in sentence selection \\
&\multicolumn{2}{|c|}{(Query, Answer)} &(position\_played\_on\_team\_speciality, power forward)\\
&\multicolumn{2}{|c|}{System Output} &point guard \\  \cmidrule[0.5pt]{2-4}
&1 &\bf{37.8}&James Patrick Johnson (born February 20 , 1987) is an American professional basketball player \\
&&& for the Toronto Raptors of the National Basketball Association ( NBA ). \\
&3 &22.9 &Johnson was the starting power forward for the Demon Deacons of Wake Forest University \\
\cmidrule[1.0pt]{2-4}
\parbox[t]{2mm}{\multirow{14}{*}{\rotatebox[origin=c]{90}{\textsc{WikiSuggest (WS)}}}}

&

\multicolumn{2}{|c|}{Error Type} & Error in answer generation \\
&\multicolumn{2}{|c|}{(Query, Answer)} &(david blaine's mother, Patrice Maureen White)\\
&\multicolumn{2}{|c|}{System Output} & Maureen\\ \cmidrule[1pt]{2-4}
&1 &14.1&David Blaine (born David Blaine White; April 4, 1973) is an American magician, illusionist $\ldots$ \\
&8 &\bf{22.6} &
Blaine was born and raised in, Brooklyn , New York the son of Patrice Maureen White
{\ldots} \\
\cmidrule[1.0pt]{2-4}
&\multicolumn{2}{|c|}{Error Type} & Noisy query \& answer  \\
&\multicolumn{2}{|c|}{(Query, Answer)} &(what are dried red grapes called, dry red wines)\\
&\multicolumn{2}{|c|}{System Output} &Chardonnay  \\ \cmidrule[0.5pt]{2-4}
&1  &         2.8     &Burgundy wine ( French : Bourgogne or vin de Bourgogne ) is wine made in the $\ldots$\\
& 2      &     \textbf{ 90.8}   & The most famous wines produced here {\ldots}
are dry red wines made from Pinot noir grapes {\ldots}
\\  \cmidrule[1.0pt]{2-4}
\multicolumn{4}{c}{\bf{Correctly Predicted Examples}}\\  \cmidrule[1.0pt]{2-4}
\parbox[t]{2mm}{\multirow{7}{*}{\rotatebox[origin=c]{90}{\textsc{WR Long}}}}&
\multicolumn{2}{|c|}{(Query, Answer)} &(position\_held, member of the National Assembly of South Africa)\\ \cmidrule[1pt]{2-4}

&1&     \bf{98.4} &   Anchen Margaretha Dreyer (born 27 March 1952) is a South African politician, a Member of  \\
&&& Parliament for the opposition Democratic Alliance , and currently $\ldots$\\
\cmidrule[1.0pt]{2-4}
&\multicolumn{2}{|c|}{(Query, Answer)}&(headquarters\_locations, Solihull)\\ \cmidrule[0.5pt]{2-4}
&1 &13.8&LaSer UK is a provider of credit and loyalty programmes , operating in the UK and Republic $\ldots$\\
&4 &\bf{82.3} &The company 's operations are in Solihull and Belfast where it employs 800 people .  \\  \cmidrule[1.5pt]{2-4}
\parbox[t]{2mm}{\multirow{3}{*}{\rotatebox[origin=c]{90}{\textsc{WS}}}}&
\multicolumn{2}{|c|}{(Query, Answer)} &(avril lavigne husband, Chad Kroeger)\\ \cmidrule[0.5pt]{2-4}
&1&17.6& Avril Ramona Lavigne ([ˈævrɪl] [ləˈviːn] / ; French pronunciation : <200b> ( [avʁil] [laviɲ] ) ;{\ldots}\\
&23&\bf{68.4}&Lavigne married Nickelback frontman , Chad Kroeger , in 2013 . Avril Ramona Lavigne was $\ldots$ \\ \cmidrule[1.5pt]{2-4}

\end{tabular}}
\vspace{-3pt}
\caption{\small{Example outputs from \textsc{Reinforce} ($K$=1) with \textsc{BoW} sentence selection model.
First column: sentence index ($l$). Second column: attention distribution $p_\theta(s_l|d,x)$. Last column: text $s_l$. }}
\vspace{-9pt}
\label{tab:full_examples} 
\end{table*}

\begin{table}
\begin{center}
\scriptsize{
\begin{tabular}{l|c|c}\toprule
& \textsc{WR Long}&\textsc{Wikisuggest}\\ \midrule 
No evidence in doc.        & 29 & 8  \\
Error in answer generation & 13 & 15\\
Noisy query \& answer       & 0 & 24 \\
Error in sentence selection & 8 & 3 \\
\toprule
\end{tabular}
}
\end{center}
\vspace{-10pt}
\caption{\small{Manual error analysis on 50 errors from the development set for \textsc{Reinforce} ($K$=1).}}
\vspace{-15pt}
\label{tab:error_categorization}
\end{table}
\paragraph{Qualitative Analysis}
We categorized the primary reasons for the errors in Table~\ref{tab:error_categorization} and present an example for each error type in Table~\ref{tab:full_examples}. All examples are from \textsc{Reinforce} with \textsc{BoW} sentence selection. The most frequent source of error for \textsc{Wikireading Long} was lack of evidence in the document. 
While the dataset 
does not contain false answers, the document does not always provide supporting evidence (examples of properties without clues are 
\textsc{elevation above sea level} and \textsc{sister}). Interestingly, the answer string can still appear in the document as in the first example in Table~\ref{tab:full_examples}: `Saint Petersburg' appears in the document (4th sentence). 
Answer generation at times failed to generate the answer even when the correct sentence was selected. This was pronounced especially in long answers. 
For the automatically collected \textsc{Wikisuggest} dataset, noisy question-answer pairs were problematic, as discussed in Section \ref{sec:data}. However, the models frequently guessed the spurious answer. We attribute higher proxy performance in sentence selection for \textsc{Wikisuggest} to noise. 
In manual analysis, sentence selection was harder in \textsc{Wikireading Long}, explaining why sampling two sentences improved performance. 

In the first correct prediction (Table~\ref{tab:full_examples}), the model generates the answer, even when it is not in the document. The second example shows when our model spots the relevant sentence without obvious clues. In the last example the model spots a sentence far from the head of the document.



\section{Related Work}
There has been substantial interest in datasets for reading comprehension. MCTest~\cite{Richardson:13} 
is a smaller-scale datasets focusing on common sense reasoning;  bAbi~\cite{Weston:15} is a synthetic dataset that captures various aspects of reasoning; and SQuAD~\cite{Rajpurkar:16,Wang:16,Xiong:16} and NewsQA~\cite{Trischler:16} are QA datasets where the answer is a span in the document. Compared to Wikireading, some datasets covers shorter passages (average 122 words for SQuAD). Cloze-style question answering datasets~\cite{Hermann:15,Onishi:16,Hill:15} assess machine comprehension but do not form questions. The recently released MS MARCO dataset~\cite{Nguyen:16} consists of query logs, web documents and crowd-sourced answers. 

Answer sentence selection is studied with the TREC QA~\cite{Voorhees:00}, WikiQA~\cite{Yang:16} and SelQA~\cite{Jurczyk:16} datasets. Recently, neural networks models~\cite{Wang:15,severyn:15,Santos:2016} achieved improvements. \newcite{Sultan:16} optimized the answer sentence extraction and the answer extraction jointly, but with gold labels for both parts. \newcite{TrischlerModel:2016} proposed a model that shares the intuition of observing inputs at multiple granularities (sentence, word), but deals with multiple choice questions. Our model considers answer sentence selection as latent and generates answer strings instead of selecting text spans. 

Hierarchical models which treats sentence selection as a latent variable have been applied text categorization~\cite{Yang:16}, extractive summarization~\cite{Cheng:16}, machine translation~\cite{Ba:14} and sentiment analysis~\cite{yessenalina2010multi,Lei:16}. To the best of our knowledge, we are the first to use the hierarchical nature of a document for QA. 

Finally, our work is related to the reinforcement learning literature. Hard and soft attention were examined in the context of caption generation~\cite{Xu:15}. Curriculum learning was investigated in~\newcite{Sachan:16}, but they focused on the ordering of training examples while we combine supervision signals. Reinforcement learning recently gained popularity in tasks such as co-reference resolution~\cite{Clark:16}, information extraction~\cite{Narasimhan:16}, semantic parsing~\cite{Andreas:16} and textual games~\cite{Narasimhan:15,He:16}.

\vspace{-4pt}\section{Conclusion}\vspace{-4pt}
We presented a coarse-to-fine framework for QA over long documents that quickly focuses on the relevant portions of a document. 
In future work we would like to deepen the use of structural clues and answer questions over multiple documents, using paragraph structure, titles, sections and more. We argue that this is necessary for developing systems that can efficiently
answer the information needs of users over large quantities of text.



\bibliography{acl2017}
\bibliographystyle{acl_natbib}

\end{document}